# The Causally Emergent Alignment Hypothesis: Causal Emergence Aligns with and Predicts Final Reward in Reinforcement Learning Agents

Federico Pigozzi[1] and Michael Levin[1,2]

[1]Allen Discovery Center, Tufts University, USA

[2]Wyss Institute for Biologically Inspired Engineering, Harvard University, USA

Author for correspondence: michael.levin@allencenter.tufts.edu

**Abstract**

A hallmark of life on Earth is the ability of agents to exert causal power and be drivers of subsequent events. This is key to cognition at all scales. Causal emergence, measuring the degree to which an agent exerts unique predictive power on its future, is one consequence of causal power. Indeed, recent discoveries have shown that biological agents, even minimal ones, increase their causal emergence after learning new memories. However, there is a major knowledge gap regarding how causally emergent artificial agents are. We focused on Reinforcement Learning (RL) of neural-network agents across an array of environmental conditions, encompassing different algorithms, agent architectures, and six environments arranged on a complexity spectrum. For consistency, we computed the causal emergence of their latent-space representations over their lifetimes. We used the recently proposed $\Phi ID$ to estimate causal emergence and tested how it related to learning performance. Our results suggested a *Causally Emergent Alignment Hypothesis*: successful agents exhibited causal emergence that was consistently predictive of final reward early in training and whose representational dynamics aligned with reward improvement in most tasks. This idea suggests that causal emergence may be a previously undisclosed axis of reorganization of neural representations in RL agents, with the potential to establish causal relationships and interventions that will lead to better RL agents. Our work also highlights the alignment between causal emergence and learning as another way biological and artificial creatures compare.

Submission type: **Full Paper**

Data/Code available at:
https://github.com/pigozzif/PhiRL

## Introduction and Related Work

A key component of living beings is their ability to act as causal agents: they exert causal power, the ability of a composite system (like a living organism made of cells and organs) to be a driver of subsequent events (Vernon et al. 2015). It is a key aspect of cognition, from brainy animals down to minimal substrates, that enables context-sensitive, goal-driven behavior in ways that distinguish them from their environment (Krakauer et al. 2020).

Causal emergence of an agent is one symptom of causal power. The causal emergence of a system measures the degree to which the whole is greater than the sum of the parts (Hoel, Albantakis, and Tononi 2013). Consider the conceptual analogy of an ant colony: if all the ants come from the same colony, they'll have a higher causal emergence than if the ants came from different colonies, because of their ability to coordinate through pheromone marks and be an emergent "self" (Watson et al. 2026).

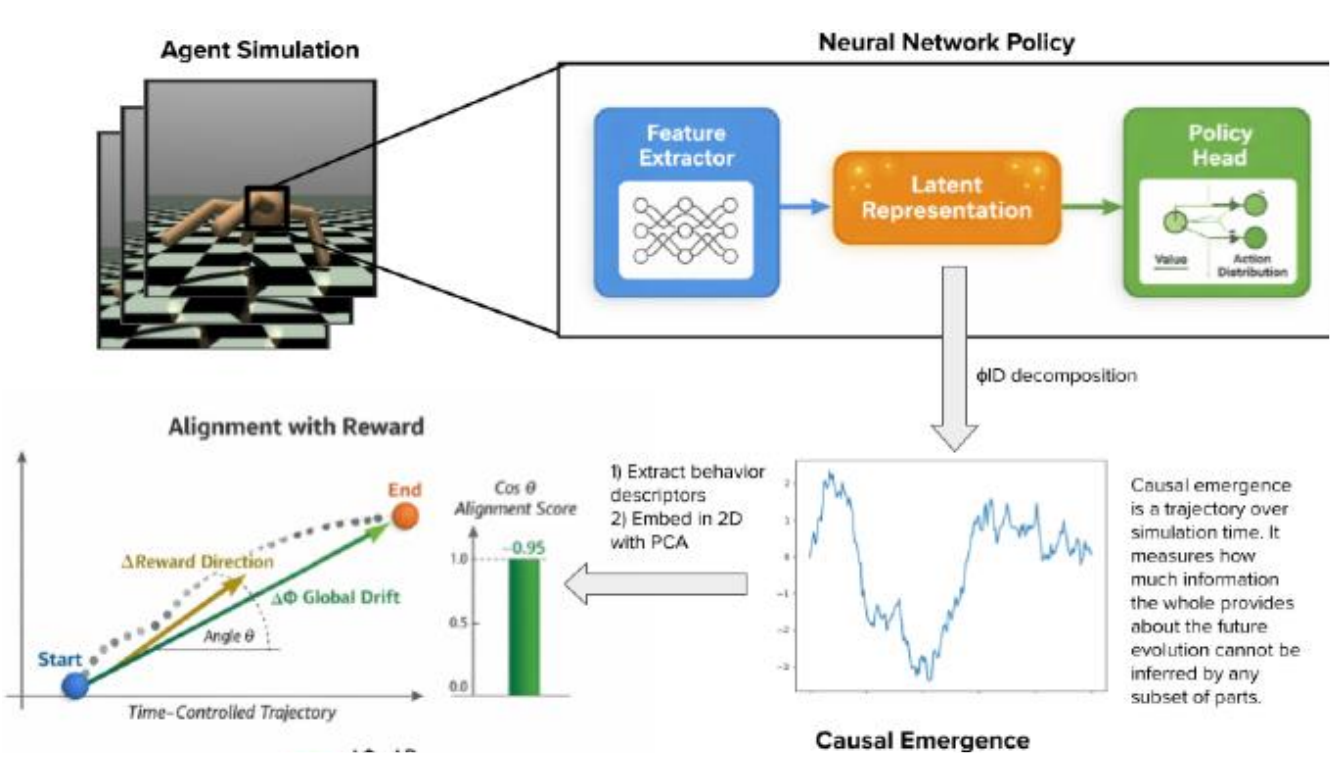

Figure 1. The schematic of our approach to computing causal emergence alignment with the reward in RL agents. Alignment is measured by whether causal emergence proceeded in the direction of increasing reward or not. We found that causal emergence had strong alignment scores across all tasks.

Among the different embodiments of causal emergence, we adopted the $\Phi ID$ decomposition (Mediano et al. 2025) because it applies specifically to multivariate time series, such as our latent-space trajectories. For a temporal system, $\Phi ID$ outputs a measure known as $\Phi^r$: expressed in natural units, the higher it is, the more causally emergent the system is (Rosas et al. 2020; Rosas et al. 2019). Different from part of the literature, we do not employ $\Phi ID$ as a hallmark of consciousness; rather, as a measure of "selfhood" or agent integration. Neuroscience studies have shown that it captures the reduction in "awareness" experienced by patients (Luppi et al. 2023).

Our recent study (Pigozzi, Goldstein, and Levin 2026) showed that $\Phi ID$ captures the reaction to learning in a minimal substrate, computational models of gene regulatory networks: indeed, these biological networks were found to increase their causal emergence after Pavlovian conditioning had induced associative memories (Biswas, Clawson, and Levin 2023). This suggested a feedback loop between agent learning and causal emergence.

This raises the question: can the same phenomenon occur in artificial agents? The field of diverse intelligence seeks to adopt tools from neuroscience and the behavioral sciences (conventionally used for brainy animals) in unconventional contexts, envisioning a continuum of cognitive abilities from active matter to conventionally intelligent agents (McMillen and Levin 2024). Artificial intelligences also sit on this continuum. Like cells or animals, they process information, learn, and pursue goals, potentially spanning the entire spectrum (Levin 2019). Reinforcement Learning (RL) agents are one notable class in this sense, since they learn from experiencing and interacting with an environment.

Previous research at the intersection of RL and causal emergence found that integrated information (measured with Tononi's Φ (Tononi et al. 2016)) correlates with fitness (Edlund et al. 2011), but limited their experiments to one task, making it hard to draw generalizable conclusions. Recent studies have also found that empowerment (a related but distinct measure of agent causal power) (Klyubin, Polani, and Nehaniv 2005) improves the efficiency of pretraining neural cellular automata (Grasso and Bongard 2023). Similar results have been found with an information bottleneck perspective (He et al. 2024). To the best of our knowledge, there has been no extensive, definitive study across multiple conditions.

We trained RL agents across a wide range of conditions, encompassing two RL algorithms, two agent architectures, and six environments distributed across a complexity spectrum (from Pendulum to Crafter). In this way, we ensured that the observed phenomena were robust to the experimental setup and reflected general properties of learning. The latent space of the neural network policy was used as a substrate because it is consistent across all environments and provides a sufficiently rich representation. If we go back to the ant colony analogy, the "parts" are the dimensions of the latent space (hidden layer neurons), while the "whole" is the latent space representation itself. We present a schematic of our pipeline in Figure 1.

Our analyses showed that causal emergence strongly aligned with the long-term (though not short-term) reward, suggesting that it provided a slow directional signal to behavior improvement. Chiefly, causal emergence from early on in training predicted final reward better than a set of standard representational metrics, meaning that it was also an early indicator of performance. Taken together, we frame these phenomena as the *Causal Emergence Alignment Hypothesis*: successful agents are those with causally emergent representations that reorganize in directions that, in most tasks, align with reward and are always predictive of final learning outcomes.

In the future, we envision these findings as a basis for revealing causal emergence as a new interventional approach for RL agents. Our results also suggest parallels between biological and artificial systems, thus highlighting how in silico experimentation complements the study of in vivo systems.

## Materials and Methods

We present the methods relevant to this study.

### Causal Emergence

*Information Decomposition*

Information theory, originally introduced to study the transmission capacity of communication channels, has emerged as a principled language for evaluating dependencies in complex systems, including artificial and biological systems. The basic object of study is Shannon's entropy:

$$H(X) = -\Sigma_x p(x) \ln p(x)$$

Where the summation is over the support of $X$, and it quantifies the amount of uncertainty about a random variable $X$. We can then define, for a process consisting of a "source" variable $X$ and a "target" variable $Y$, the mutual information as the uncertainty that is left on $Y$ after observing $X$, i.e., how much information observing $X$ discloses about $Y$.
But what if there is more than one source variable, as in complex systems like RL neural network policies? We must then consider all the different directions in which information can flow in a system. Intuitively, let us consider the case of stereoscopic vision in humans: with one eye open, we perceive a unique set of visual features for each eye, as well as redundant features captured by both eyes. Depth perception, which can only be captured if both eyes are open simultaneously, corresponds instead to synergistic information. The seminal work on Partial Information Decomposition (PID[1]) provides a framework for partitioning mutual information into these three information atoms (redundant, unique, and synergistic).

[1] Williams, P.L. and Beer, R.D., 2010. Nonnegative decomposition of multivariate information. *arXiv preprint arXiv:1004.2515*.

*Causal Emergence*

Our latent space trajectories are multivariate, consisting of the activations of several neurons. We relied on the recent framework of Integrated Information Decomposition $\Phi ID$, which is the PID's extension to multivariate data (Mediano et al. 2025). According to general assumptions outlined in (Rosas et al. 2020), a system's capacity to display emergence depends on how much information the whole provides about the future evolution that cannot be inferred by any subset of parts. The $\Phi ID$ formal apparatus then tells us that we can decompose this capacity as the sum of two terms:
1) *Downward causation*: the amount of information that the whole predicts about the future of the single components.
2) *Synergy*: the amount of information that the whole predicts about the future of the whole.
We illustrate the definition in Figure 2 for a generic system made of $n$ parts. This definition previously appeared to quantify the reduction in emergence capacity between healthy and brain-injured patients (Luppi et al. 2023) and reaction to learning in simulated gene regulatory networks (Pigozzi, Goldstein, and Levin 2026); we chose it as our measure of causal emergence because it accounts for all types of influences a system can have on its future.

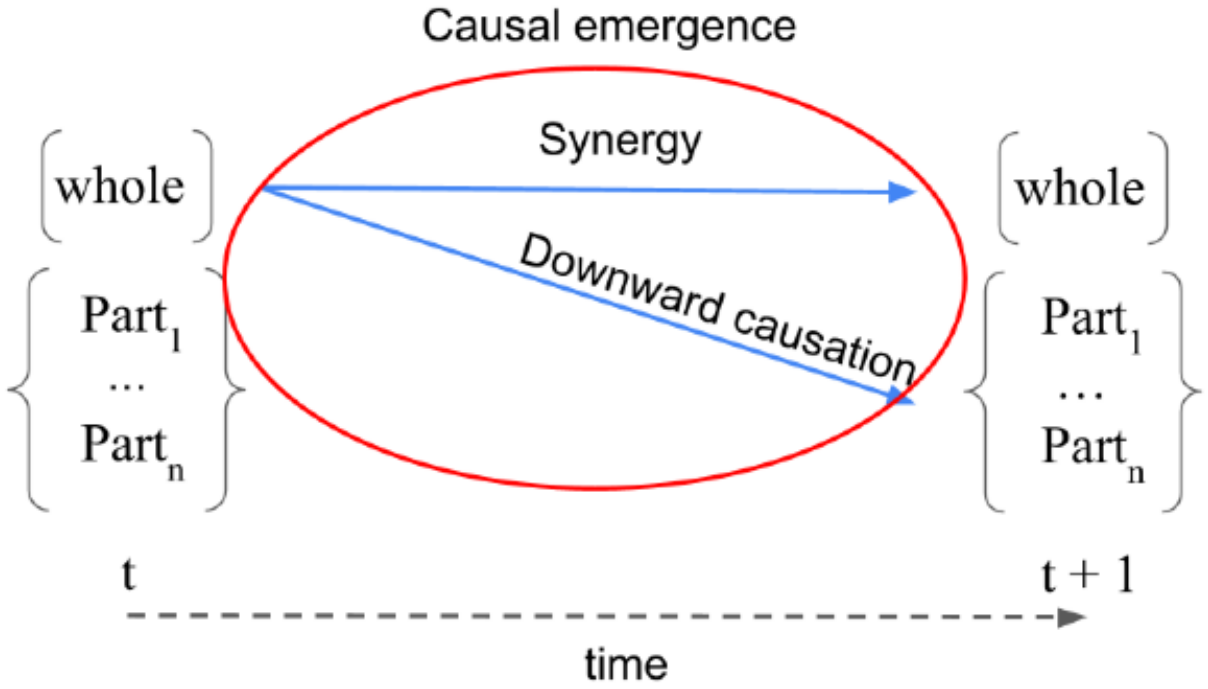

Figure 2. Causal emergence is the sum of the amount of information that the whole predicts about the future of the single components (synergy) and the amount of information that the whole predicts about the future of the whole (causal decoupling).

Other measures of agent integration exist, such as total correlation and co-information. But, they are instantaneous measures; they fail to capture the temporal and causal aspects of information dependencies up to and including all future time steps, a crucial aspect for dynamical systems that evolve over time (Mediano et al. 2025), such as our latent-space trajectories.

*Gaussian Information Theory*

Information theory was originally defined for discrete random variables, but our representations, being neural network activations, are continuous-valued. Hence, we used the continuous generalization of Shannon's entropy, the differential entropy:

$$H(X) = \int_x p(x) \ln p(x)\, dx$$

This integral is generally hard to compute because it requires estimating $p(x)$. But if we assume that $p(x)$ follows a Gaussian distribution, we can leverage closed-form estimators for the entropy and, as a result, all the other information measures (Barrett 2015). Indeed, the bivariate mutual information (in natural units) becomes:

$$I(X,Y) = -\frac{(1-\rho^2)}{2}$$

Where $\rho$ is the Pearson correlation coefficient between $X$ and $Y$.
Most practical computations of causal emergence converge on the same simple form for Gaussian continuous variables that we adopted here. We first computed the lag-1 mutual information matrix for all node pairs in the system using the equation above (Blackiston et al. 2025). Since we cannot handle systems with many parts due to the combinatorial complexity (Kitazono, Kanai, and Oizumi 2018), we reduced the dimensionality using the minimum-information bipartition (Toker and Sommer 2019). This bipartition bisects the system into two components by approximating the bisection through the Fiedler vector (the eigenvector of the graph Laplacian corresponding to the smallest non-zero eigenvalue). After bisecting the graph with the Fiedler vector, we averaged within each component and compared the dynamics of the two parts to the whole. In essence, we sliced a watermelon along its longest axis and measured how well the average number of seeds in one half predicted the average number of seeds in the other half. Finally, we solved a linear system of equations relating the mutual information to the atoms in which the $\Phi ID$ is decomposed, including the downward causation and causal decoupling; their sum is our measure of causal emergence.

## Reinforcement Learning and $\Phi^r$ substrate

RL (Sutton and Barto 2018) provides a framework for training an agent to solve a task through trial and error, akin to how animals learn in the real world (Neftci and Averbeck 2019). Agents are neural network policies that receive the state of the world as input and output the agent's action for that time step. A reward signal from the environment measures the quality of the action taken and is used to update the policy's weights and biases, enabling the agent to successfully learn a task.

We adopted the standard agent architecture from the literature. At each time step $t$, a feature extractor $f$ maps the input state $s_t$ to a latent representation $f(s_t) = z_t \in \mathbb{R}^{d_{\text{latent}}}$. $z_t$ is then fed to a policy head π to obtain the action $a_t = \pi(z_t)$. We used stable-baselines3 as our RL framework (Raffin et al. 2021).

Causal emergence computes the agent integration within a system, but what counts as the system in our case? We wanted a system representation that is:

- Low-dimensional enough to make the $\Phi ID$ computation tractable and not noisy;
- Dynamically rich enough to show meaningful variation;
- Consistent across environments and architectures.

With these desiderata in mind, we found the latent $z_t$ to be the best candidate. Indeed, after preliminary experiments and relying on canonical values in the literature, we found $d_{\text{latent}} = 64$ to provide sufficient representational capacity without making the $\Phi ID$ estimation noisy, while remaining consistent across all experimental conditions because of the architecture definition (the action distribution, for example, changes according to the environment).

In the following, we computed causal emergence on the latent representation trajectories $Z = [z_0, z_1, \dots, z_T]$ for an agent episode that lasted $T$ time steps (before being truncated or terminated according to the environment's meaning).

## Data Preprocessing

Gaussian Information Theory assumes that the variables are jointly Gaussian. Since this is hardly ever true for neural activations, we applied a copula-based Gaussianization (rank-normal transform) to ensure approximate marginal normality. After this transformation, only a minority of all units (28.53%) rejected the normality hypothesis (D'Agostino $K^2$ test, $p<0.05$). Subsequently, we z-scored the data to standardize it (Blackiston et al. 2025; Pigozzi, Goldstein, and Levin 2026).

## Environments

We ran all the experimental combinations across six OpenAI Gymnasium[2] environments (except CrafterReward, which comes in its own library[3]), as shown in Figure 3. Our environments spanned a complexity spectrum, from least to most complex:

- Pendulum-v1: The agent applies the torque to keep the pole in an upright position. It is a classic task with minimal state and action dimensions.
- Lunar-Lander-v2: The agent turns the engines on or off to land as close as possible to a landing pad. It is a more complex control task, with more actions and states.
- BipedalWalker-v4: The agent controls a bipedal 2D robot to walk upright as far as possible. It has a high-dimensional state space.
- Walker2D-v4: The agent controls a bipedal robot 3D consisting of seven body parts. It is embodied, with continuous states, actions, and non-trivial body dynamics.
- Ant-v4: The agent controls a four-legged 3D ant consisting of nine body parts. It is embodied, with continuous states, actions, and non-trivial body dynamics.
- CrafterReward-v1: In the 2D version of Minecraft, the agent must survive as long as possible while gathering resources and fending off enemies. It requires long-term planning and exploration skills.

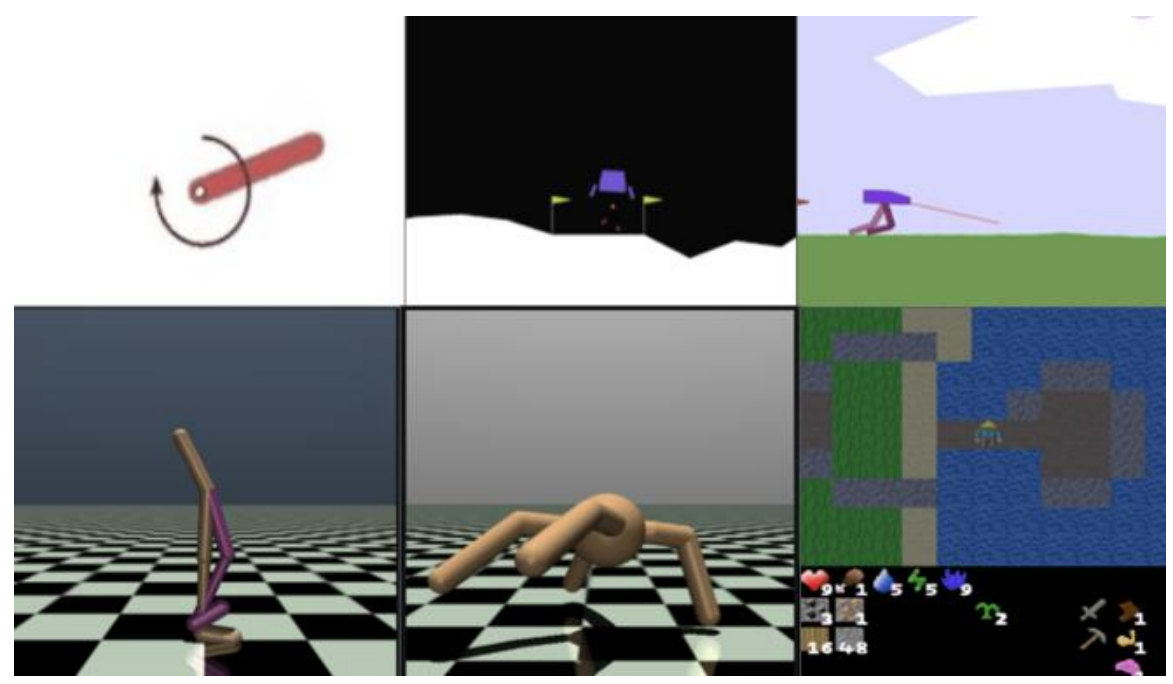

Figure 3: The six environments. From top to bottom, left to right: Pendulum, LunarLander, BipedalWalker, Walker2D, Ant, and CrafterReward.

## Architectures and Algorithms

To experiment with a broad range of conditions, we tested two architectures and two RL algorithms across all environments. As architectures, we used a feed-forward (MLP) and a recurrent neural network (GRU) because they impose different inductive biases on representations (static vs. temporally dependent), allowing us to isolate the effect of architectural inductive biases (memory vs. non-memory).

We implemented the MLP using the default stable-baselines3 policy architecture. For the sake of fairness, we implemented the GRU variant using the same architecture as the MLP, but with a GRU block added on top of the feature extractor. This resulted in slightly different parameter counts for the two variants, but our focus was how causal emergence behaved across different architectural classes (memory vs. memoryless) rather than performance or capacity equivalence. We also remark that for CrafterReward, whose state space consists of images, we replaced the feed-forward feature extractor with the same convolutional backbone used in (Mnih et al. 2015), as is usually done for such environments. For all architectures, we coded their blocks in PyTorch using the library's default parameters.

[2] https://github.com/farama-foundation/gymnasium

[3] https://github.com/danijar/crafter

As algorithms, we employed Proximal Policy Optimization (PPO[4]) and Soft Actor-Critic (SAC) (Haarnoja et al. 2018), both of which are among the most established methods from two significantly different families (on-policy vs. off-policy; deterministic vs. entropy-regularized). In this way, we verified if the observed phenomena were robust to the specific algorithm. We remark that for CrafterReward and LunarLander, which have discrete rather than continuous actions, we replaced SAC with Deep Q-Network (DQN) (Mnih et al. 2015), the established off-policy method for these spaces.

## Experimental Parameters

For each environment × algorithm × architecture combination, we performed 10 runs with different random seeds for reproducibility. Every $5 \times 10^3$ time steps, we froze the policy's weights and biases and checkpointed its activations across 10 test episodes (with different random seeds) to reduce noise and chiefly, ensure statistically reliable estimates of causal emergence. In total, this yielded $6 \times 2 \times 2 \times 10 \times 21 \times 10 = 50{,}400$ episodes.

For each episode, we computed the causal emergence trajectory over the simulation's history of latent-space activations and aggregated it with the median to have a robust summary, as previously done in (Pigozzi, Goldstein, and Levin 2026).

Each run lasted $10^6$ environment steps; we found this number sufficient for most runs to solve the tasks and plateau. All other parameters were kept at their default stable-baselines3 values to isolate the representation effect from tuning and report the results for off-the-shelf, widely used settings.

We coded all experiments and analyses in Python and made them publicly available at https://github.com/pigozzif/PhiRL.

## Results

To understand how causal emergence reacted to learning, we answered the following Research Questions (RQ):

- RQ0: Does causal emergence capture novel information? In other words, does it overlap with known representation metrics?
- RQ1: Does causal mergence align and move with the reward?
- RQ2: Does causal emergence predict learning outcomes?

This transition gradually deepened our task from descriptive to functional, and then to predictive.

### RQ0: Did causal emergence correlate with known representation metrics?

To ensure our causal emergence approach captured a novel axis of representation shift, we tested whether it was uncorrelated with other established neural representation metrics. Our metrics were standard measures from information theory and dynamical systems: entropy, Shannon mutual information, autocorrelation, effective dimension, and magnitude of the latent representations.

For each run of every experimental combination, we computed Spearman's ρ rho between each metric and causal emergence. We reported the results in Figure 4 as the % of significant ($p<0.05$) runs for each experimental combination. As suggested by the deep purple color of the array, no environment-metric pair had a significant fraction higher than 6%, with the vast majority hovering around 0%. These numbers meant that causal emergence did not merely co-fluctuate with other underlying variables. Contrarily, it captured a novel axis, justifying our next experiments.

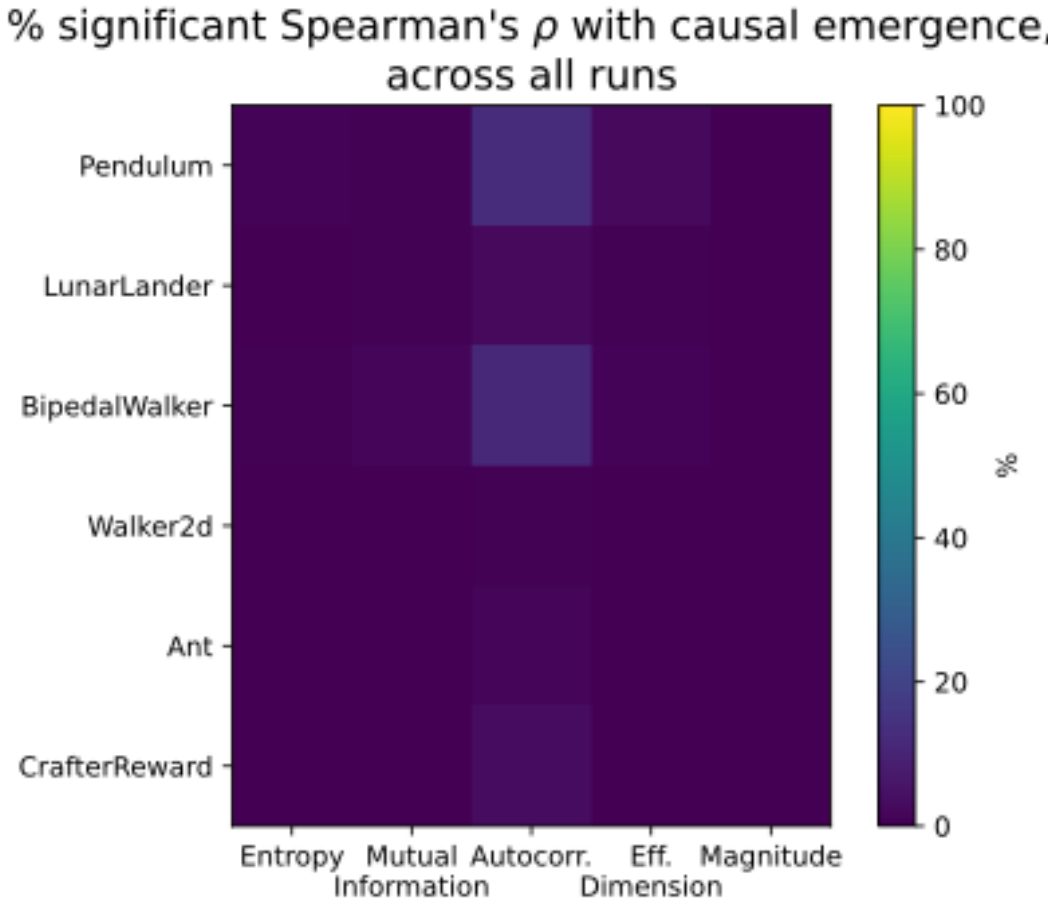

Figure 4: % of runs that had a significant ($p<0.05$) Spearman's ρ between standard neural representation metrics (x-axis labels) and causal emergence, for each

[4] Schulman, J., Wolski, F., Dhariwal, P., Radford, A. and Klimov, O., 2017. Proximal policy optimization algorithms. *arXiv preprint arXiv:1707.06347*.

environment (y-axis). Deep purple dominates the array, with a maximum value of only 6% for Autocorrelation+Pendulum, meaning that causal emergence did not generally correlate with established representation metrics of neural latents.

## RQ1: Did causal emergence align with the reward?

Having established that causal emergence did not overlap with existing representation metrics, we investigated whether its changes were oriented toward improving performance. Concretely, we computed the *reward alignment* of a trajectory $\tau \in \mathbb{R}^T$ (e.g., causal emergence or other representation metrics) and reward signal $r \in \mathbb{R}^T$, where $T$ is the length of the RL episode, as follows:

1) Embed the trajectory using PCA to have a low-dimensional embedding $e \in \mathbb{R}^m$;
2) Fit a linear model with coefficients $w \in \mathbb{R}^T$ to predict $r$ from $e$;
3) Interpret the coefficients $w$ as a reward gradient in the embedding space;
4) Compute the cosine similarity between $w$ and the global direction (i.e., end minus start, for *global* reward alignment) or the mean of the instantaneous directions (for *local* reward alignment):

$$Global\ Reward\ Alignment = \frac{\langle w, e_T - e_0 \rangle}{||w||\,||e_T - e_0||}$$

$$Local\ Reward\ Alignment = \frac{1}{(T-1)} \sum_{t=0}^{T-1} \frac{\langle w, e_{t+1} - e_t \rangle}{||w||\,||e_{t+1} - e_t||}$$

Both scores took values in $[-1,1]$. A high alignment meant that the representation metric under consideration proceeded in the direction of increasing reward, linking representational dynamics to functional improvement.

Reward alignment measured directional consistency between representational change and performance improvement. Other measures, such as statistical correlations and regression coefficients, capture association but not the trajectory over time. Similarly, distance metrics ignore whether the change is useful for reward. In other words, reward alignment evaluated if the metric's path was effectively goal-directed and linked to behavior.

To embed a causal emergence trajectory, we described it with seven "behavior" descriptors that were found to be the most expressive and compact at the same time (Pigozzi, Goldstein, and Levin 2026):

1) *Standard deviation*: a measure of dispersion.
2) *Trend*: the slope of the least squares fit of the trajectory. A positive slope indicated an increasing trend, while a negative slope indicated a decreasing trend.
3) *Monotonicity*: the Kendall's tau coefficient between the trajectory and the sequence of its time stamps. Kendall's tau is a standard statistic to measure ordinal association between two quantities, and in our case, it was the highest for a perfectly monotonically increasing trajectory, and the lowest for a perfectly monotonically decreasing one, with values around zero corresponding to the trajectory fluctuating independently of the time axis.
4) *Flatness*: how flat the trajectory was and did not locally deviate from the mean. We divided the trajectory into consecutive intervals and approximated it with the mean of each interval. We computed flatness as the R-squared coefficient of this approximation: the higher the coefficient, the better the fit of the local means, indicating that the trajectory was locally flat (though jumps may have occurred at the interval boundaries). After preliminary experiments, we found an interval size of 100 to be sufficient to capture the intuition behind a trajectory's flatness.
5) *Number of peaks*: the number of local minima and maxima of the trajectory.
6) *Average distance among peaks*: the average distance among all the peaks from 4) (or 0 if there were none).
7) *Average difference among peaks*: the average difference in causal emergence value of the peaks from 4) (or 0 if there were none).
8) *Range*: the difference in causal emergence value between the maximum and the minimum peaks.

For step 1), we tested $m = 2,3,\ldots,8$ for the PCA embedding dimension and found the results to be robust; in the following, we reported them with $m = 2$ for simplicity. Before fitting the linear model, we residualized both the embedding and the reward with respect to time by regressing each on time (i.e., time step index) and using the residuals for subsequent analyses. In this way, we excluded the confounding that both may have been trivially drifting together over time. We found the results without time residualization to be comparable to those with time residualization, demonstrating their robustness; in the following, we reported results with time residualization for completeness.

Table 1 shows the global and local alignment scores for causal emergence. The global alignment scores were of large magnitude, indicating that causal emergence aligned strongly with the same (or opposite) direction as reward improvement. Also, the sign was positive in a majority (5/6) of environments, meaning the same direction, whereas negative in CrafterReward, which may be linked to more time spent on early exploration for that task. Contrarily, the local alignment scores were approximately zero. This meant that causal emergence captured the agent's long-term “representation shift” or behavioral adaptation, but not the step-by-step improvements in reward, which were noisy or canceled each other out.

| | Global Reward Alignment | Local Reward Alignment |
|---|---|---|
| Pendulum | 0.99 | 0.00 |
| LunarLander | 1.00 | 0.00 |
| BipedalWalker | 0.86 | 0.00 |
| Walker2D | 0.35 | 0.03 |
| Ant | 0.49 | 0.02 |
| CrafterReward | -0.95 | 0.00 |

Table 1: global and local reward alignment scores for causal emergence, across all six environments. Reward alignment measured the degree to which the trajectory in embedding space aligned with the direction that increased reward; global was for the whole trajectory, while local was the mean of the instantaneous angles. Global alignment was strong in magnitude, while the local alignment was negligible. While short-term changes (local) were unrelated to reward, long-term changes (global) were strongly aligned with reward, indicating a slow representational drift relating to learning.

We found the global alignment scores of causal emergence to be significantly greater than those obtained with random projections (Mann-Whitney U test, $p<0.05$). At the same time, the scores were not significantly different from those of the standard representational metrics for RQ0. It meant that causal emergence did not capture a quantitatively different direction but instead summarized information scattered across many (weaker and heterogeneous) signals. We verified whether the reward alignment changed with the architecture and the algorithm. We found no significant differences in scores between MLP and GRU, and between PPO and SAC/DQN, thus excluding effects related to the experimental configuration.

## RQ2: Did causal emergence predict the final reward performance, and how did it compare to the baselines?

Correlation alone does not prove if a measure is relevant for learning. Hence, we moved to a predictive setting, asking whether early measurements of these predictors foreshadowed final performance. This reframed the question to understand if it captured information that mattered for learning outcomes.

To answer this question, we trained a machine learning model to predict the reward at the last time step (taken as the median over the last 10 test episodes), using as input the representation metrics from the first $2 \times 10^5$ time steps (i.e., the first 20% of each run). As a machine learning model, we used random forest, since it is not sensitive to hyperparameter choices, but we found other methods (linear regression and MLPs) to achieve qualitatively similar results. We tested causal emergence as input against the representation metrics from RQ0 (as baselines). Prediction scores were obtained using 5-fold cross-validation.

Figure 5 shows the results, one plot per environment, with Spearman’s $\rho$ between the ground-truth and the predicted values (the higher, the better the prediction performance). The baseline boxes report the performance for the best predictor among the baselines. We found the differences between causal emergence and the baselines to be statistically significant across all environments (Mann-Whitney U test, $p<0.05$). It meant that, across all considered environments, causal emergence was a better predictor of the final learning outcome than established representation metrics.

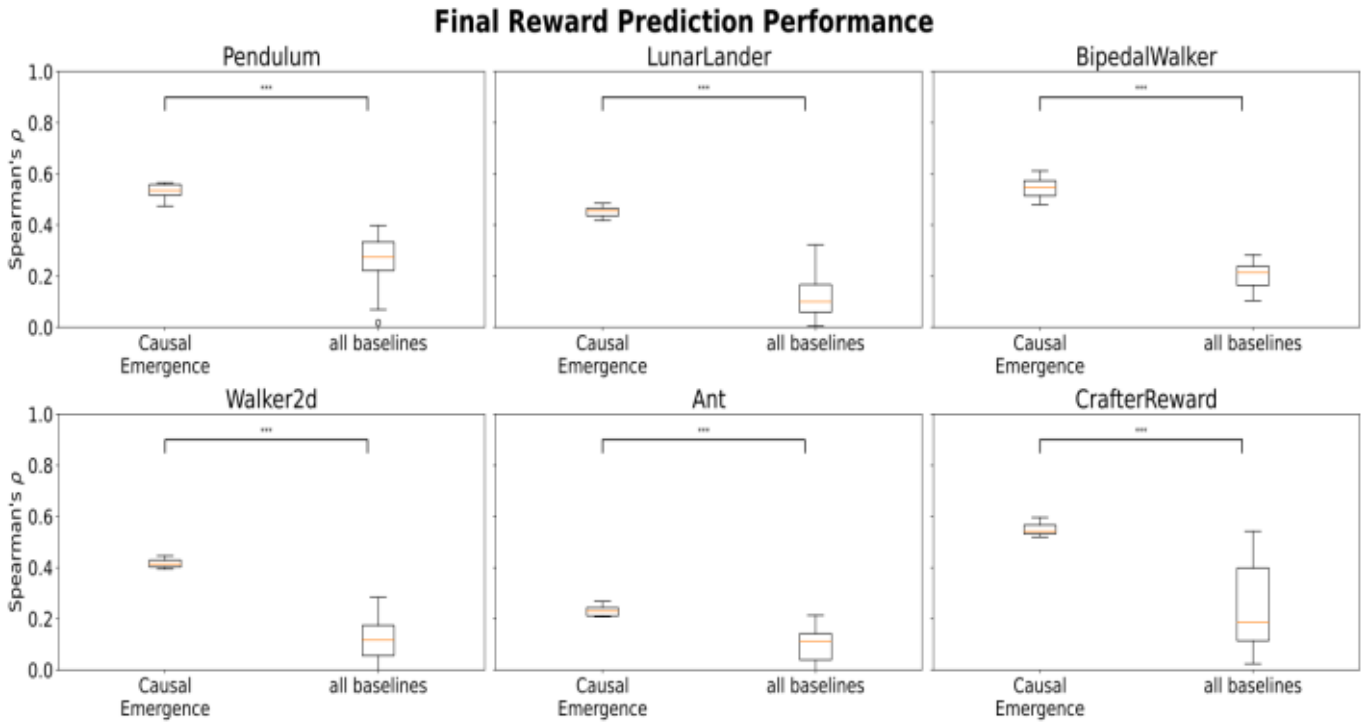

Figure 5: Prediction performance of a random forest regressor, measured as Spearman’s $\rho$ between the ground-truth and the predicted values. For each run, the regressor was trained to predict the reward at the final time step, using the metric on the x-axis (causal emergence descriptors or baseline representation metrics) as input. For the baseline boxes, we reported the best performance across all metrics. In all environments, causal emergence achieved superior prediction performance.

To push our analysis one step further, we repeated the experiment, using all baselines together as predictors for the same model (rather than one at a time and reporting the best), and then repeated the experiment with the addition of causal emergence descriptors. We reported the results in Figure 6, using the same syntax as Figure 5.

The plots and the significance tests (with Mann-Whitney) revealed that this time, causal emergence was worse in 4/6 environments, the same in 1/6, and better in 1/6. Conversely, adding causal emergence to the baselines did not worsen performance in 3/6 environments and improved it in the other 3.

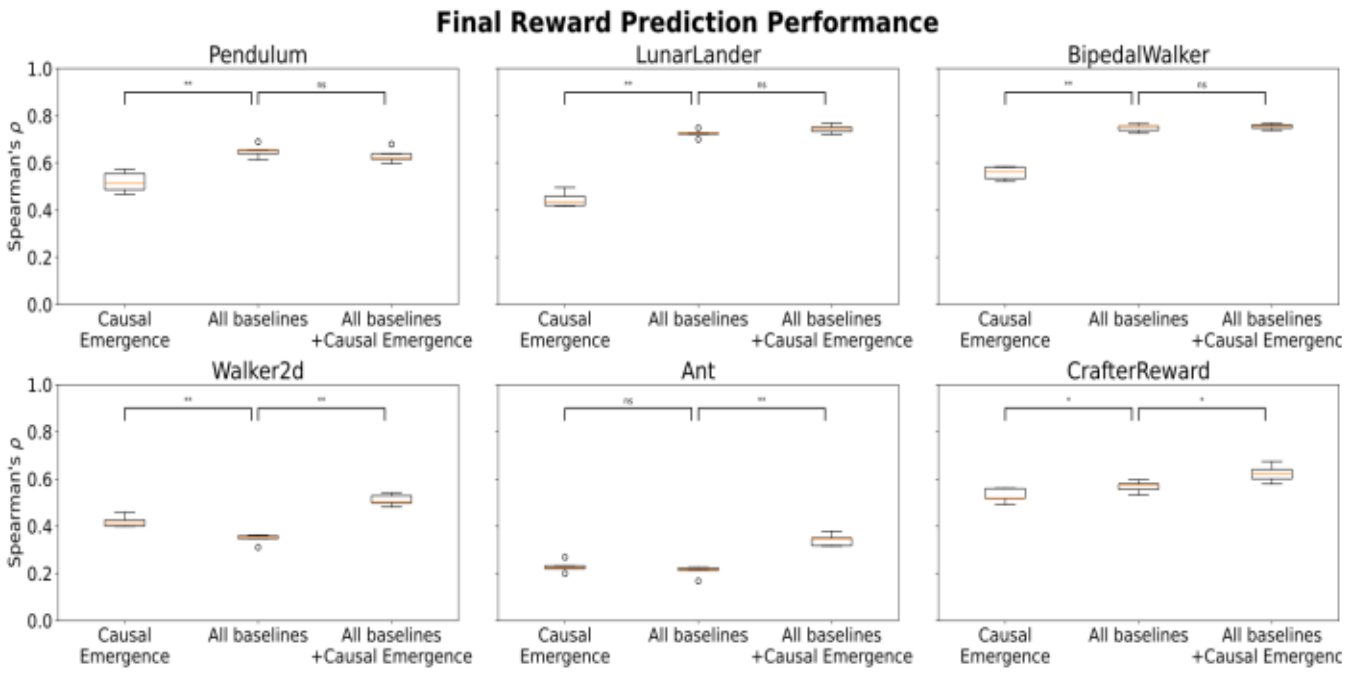

Figure 6: Prediction performance of a random forest regressor, measured as Spearman's $\rho$ between the ground-truth and the predicted values. For each run, the regressor was trained to predict the reward at the final time step, given the measure on the x-axis (causal emergence descriptors, all the baselines, or both) as input. Causal emergence alone did not outperform the combined baseline metrics. But adding it to them significantly increased performance in 3/6 environments and did not worsen performance in the other 3/6.

## Discussion

Causal emergence trajectories had very strong global but near-zero local reward alignment across all environments. Reward alignment captured the degree to which the changes in the representation metric aligned with the direction that improved learning. This suggested that learning induced a slow, long-term representational reorganization that was captured by causal emergence, whereas local changes were noisy or canceled out. Complementing this result, we found that causal emergence was a better predictor of final learning performance than standard representational metrics like entropy, Shannon mutual information, autocorrelation, effective dimension, and magnitude. This meant that the captured aspects of representation reorganization were functionally relevant for downstream learning, beyond standard metrics. Together, these results suggested that causal emergence provided a directional signal and an early indicator of final performance.

Notably, causal emergence was a better predictor when the baselines were considered separately. This suggested that causal emergence was not "the best predictor of all," but rather a low-dimensional summary of several representational metrics; in other words, it did not redundantly overlap with them but instead compressed distributed, weaker signals into a single geometric object, in line with recent discoveries for manifold learning (Varley et al. 2025).

We concluded that successful RL agents have causally emergent representations that reorganize in directions that align with reward and are always predictive of final learning outcomes. This is consistent with a prior proposal in which an agent's border is determined by the size of its goals; like cells adapting to reach homeostasis, RL agents reach a preferred state and establish selves, which is what causal emergence may be capturing (Levin 2019). But for the moment, our results cannot establish causality or whether causal emergence drives learning progress.

In the future, we will investigate whether predicting learning also implies causality; in other words, if intervention in the causal emergence space affects learning in a directed manner. If this held true, causal emergence would potentially spur advances in RL algorithms. We also look forward to testing not only the causal emergence *of* an agent, but also the interactions *between* the agent and its environment, possibly drawing on the active inference (Fields, Goldstein, and Sandved-Smith 2024), curiosity search (Gottlieb et al. 2013), and skill discovery communities (Etcheverry et al. 2025).

In addition to these experiments, we will address the main limitation of this work: that any conclusions we draw are limited by the array of experimental conditions we considered. We will extend our analyses to more complex architectures, like world models (Ha and Schmidhuber 2018; Gao et al. 2025). As for the environments, we will also investigate more complex embodied (Savva et al. 2019)and procedurally generated tasks (Mohanty et al. 2020). The latter will allow us to test the intriguing hypothesis that causal emergence predicts, or is causally related to, generalization to novel environments, an all-important topic in RL (Ghosh et al. 2021). Finally, it will be intriguing to link our findings to the broader theory of representations in deep learning, such as the information bottleneck principle (Tishby and Zaslavsky 2015).

When considered in the context of the biological literature, our results demonstrate a new parallel between biological and artificial creatures. Indeed, biological substrates such as gene regulatory networks have been found to exhibit similar causal emergence in response to learning (Pigozzi, Goldstein, and Levin 2026). To the extent that significant symmetries link all agents, regardless of their composition, origin story, scale, or problem space (Levin 2022; Clawson and Levin 2023; Tok, Powell, and Guellaï 2025), studies using tools from computational neuroscience in synthetic models are poised to advance the fields of diverse intelligence, AI, cognitive science, evolutionary biology, and engineering.

## Acknowledgements

We thank Patrick McMillen for the ant colony analogy to explain causal emergence, and Astonishing Labs for their support via a sponsored research agreement.